\documentclass[letterpaper,times]{sae}

\usepackage[svgnames]{xcolor}
\definecolor{SAEblue}{RGB}{1,160,233}
\raggedright

\usepackage[justification=justified, singlelinecheck=false]{caption}  
\DeclareCaptionFont{eightpt}{\fontsize{8pt}{11pt}\selectfont #1}
\usepackage{lineno}
\usepackage[utf8]{inputenc}

\usepackage{array}
\newcolumntype{L}[1]{>{\raggedright\let\newline\\\arraybackslash\hspace{0pt}}p{#1}}
\newcolumntype{C}[1]{>{\centering\let\newline\\\arraybackslash\hspace{0pt}}p{#1}}
\newcolumntype{R}[1]{>{\raggedleft\let\newline\\\arraybackslash\hspace{0pt}}p{#1}}

\usepackage{graphicx}
\usepackage{multirow}
\usepackage{amsmath}

\newcommand{\ignore}[1]{}

\usepackage{nameref}
\usepackage{booktabs}
\usepackage{color,soul}

\makeatletter
\def\@seccntformat#1{%
  \expandafter\csname c@#1\endcsname\c@section
  }
\makeatother

\usepackage{subfigure}
\usepackage{multimedia}

\PaperTitle {Active Collision Avoidance System for E-Scooters in Pedestrian Environment}

\makeatletter 
\renewcommand\@biblabel[1]{#1. } 
\makeatother

\AddAuthor{Xuke Yan $^*$}{Department of Computer Science and Engineering, Oakland University, Rochester, MI, USA}
\AddAuthor{Dan Shen $^*$}{Department of Electrical and Computer Engineering, Purdue University, West Lafayette, IN, USA}

\PaperNumber{20xx-XX-XXXX}

\begin{document}
\maketitle

\def\thefootnote{*}\footnotetext{These authors contributed equally to this work}

\section{Abstract}
In the dense fabric of urban areas, electric scooters have rapidly become a preferred mode of transportation. As they cater to modern mobility demands, they present significant safety challenges, especially when interacting with pedestrians. In general, e-scooters are suggested to be ridden in bike lanes/sidewalks or share the road with cars at the maximum speed of about 15-20 mph, which is more flexible and much faster than pedestrians and bicyclists. Accurate prediction of pedestrian movement, coupled with assistant motion control of scooters, is essential in minimizing collision risks and seamlessly integrating scooters in areas dense with pedestrians.
Addressing these safety concerns, our research introduces a novel e-Scooter collision avoidance system (eCAS) with a method for predicting pedestrian trajectories, employing an advanced LSTM network integrated with a state refinement module. This method predicts future trajectories by considering not just past pedestrian positions but also accounting for the behavior and locations of surrounding individuals, acknowledging the influence of human interactions.
Leveraging the pedestrians’ estimated trajectories based on their historical behaviors, we have devised an e-scooter path planning system that relies on an interpolating curve planner, which can continuously analyze the driving scene, understand the behavior of other road users, evaluate the risk assessment, and predict its future trajectory. This proactive model is designed to ensure unobstructed movement in areas with substantial pedestrian traffic without collisions.
Results are validated on two public datasets, ETH and UCY, providing encouraging outcomes. Our model demonstrated proficiency in anticipating pedestrian paths and augmented scooter path planning, allowing for heightened adaptability in densely populated locales. This study shows the potential of melding pedestrian trajectory prediction with scooter motion planning. With the ubiquity of electric scooters in urban environments, such advancements have become crucial to safeguard all participants in urban transit.
\section{1. Introduction} \label{sec:introduction}

Electric scooters (e-scooters) have seen a rapid rise in adoption in cities worldwide over recent years \cite{hardt2019usage}. Their compactness, efficiency, and ability to swiftly traverse urban terrain make them a sought-after transportation choice for short-distance travel. As cities grapple with increasing congestion and pollution challenges, e-scooters offer an eco-friendly solution, reducing the number of cars on the road while catering to the convenience-centric lifestyles of modern urban dwellers \cite{moreau2020dockless}. However, the very attributes that make e-scooters popular – their agility and speed – also present new safety complexities, particularly when navigating pedestrian-dense environments \cite{heydari2022investigating}. Although there are designated lanes in some urban areas for bikes and scooters, they often find themselves intermingling with pedestrians, either due to the lack of dedicated infrastructure or user preference \cite{gossling2020integrating}. When an e-scooter with the ability to travel up to 20 mph collides with a pedestrian, the outcome can be grave  \cite{yang2020safety}. Furthermore, the behavior of pedestrians can be quite unpredictable. Sudden stops, changes in direction, or clustering in groups can pose immediate challenges to e-scooter riders aiming to avoid collisions \cite{tuncer2020scooters}.

Current e-scooter safety mechanisms often involve passive safety measures, such as reflectors, lights, and bells. While these features are essential for basic safety, they are reactive in nature. The e-scooter rider must constantly be vigilant, reacting in real-time to potential obstacles or risky situations. What is missing from this paradigm is a proactive approach that can anticipate and navigate through potential collision scenarios before they pose immediate threats \cite{ma2021scooter}. In the realms of autonomous vehicles and robotics, research has made significant strides in trajectory prediction and path planning \cite{hubmann2018automated, katrakazas2015real}. Predictive models enable these vehicles to anticipate the movement of other road users and make informed decisions, thereby increasing safety. E-scooters, by virtue of their smaller form factor and unique operating scenarios, require tailored solutions that cater to their specific challenges.

Automated vehicles (AVs) are more popular and
attractive with the tremendous and impressive technological progress in recent years. Noticeable achievements in both software and hardware have brought more automation functionality into reality, which reduces severe and fatal crashes that are related to vulnerable road users (VRU) \cite{shen2019test}, and research topics in VRU safety are an indispensable part of the advent of AVs driving in cities. The collision avoidance system is one of the AV active safety features for handling road departure VRU-related collision problems, which has been devised and equipped on some production vehicles \cite{shen2020collision}. Although many efforts have been put into the data set development \cite{prabu2022wearable}, e-Scooter risk assessment \cite{prabu2022risk}, and sensor fusion technique \cite{shen2023fusion}, to the best of the authors' knowledge, there are not many research developments and progress on collision avoidance for e-Scooters with vehicles, pedestrians, and bicyclists. This situation is more significant with the increased popularity of micro-mobility vehicles like e-scooters. These tools tend to be small, fast, agile, and share the road with both cars and VRU.

With this backdrop, our research seeks to bridge this gap, bringing the scooter collision mitigation system with advanced trajectory prediction and path planning to e-scooters. The proposed system employs state-of-the-art methodologies, aiming to enhance the safety of e-scooter riders and pedestrians alike, while ensuring fluidity in movement even in crowded urban environments. Through this paper, we shed light on our methodologies, validate our approach using renowned datasets, and elucidate the potential of integrating proactive safety measures in e-scooters, ensuring safer urban transit for all. Due to page limitation, this paper only introduces the scooter-based collision mitigation system until the motion planner. The eCAS integrated with the control part will be celebrated in future publications.

This work presents the following contributions:
\begin{itemize}
\item \textbf{e-Scooter Collision Avoidance System:} Proposed a new e-Scooter collision avoidance system with the basic concept, overall architecture with pedestrian trajectory prediction and motion planner, operation process, and experiment results. The new system can be utilized in the environment with both e-scooters and pedestrians.

\item \textbf{Pedestrian Trajectory Prediction Module:} The proposed eCAS system is enhanced with a pedestrian trajectory prediction module that employs an LSTM network, accounting for the dynamics of urban pedestrian movement. This integration enhanced proactive safety adjustments in e-scooter navigation within dense areas.

\item \textbf{E-Scooter Path Planning Module:} 
The dynamic traffic environment for e-Scooter was modeled using the artificial potential field, which is able to identify risk levels for all moving trajectories. The gradient decent-based motion planner was able to generate the collision avoidance path via the energy surface of the potential.
\end{itemize}

These contributions, taken together, offer valuable insights and practical tools for improving the safety and efficiency of e-scooters in urban environments dense with pedestrian traffic.

This work is organized as follows. In Section 2, we conduct a review of existing literature, highlighting works on pedestrian trajectory prediction and vehicle path planning. In Section 3, we present the concept and operation process of the proposed eCAS with pedestrians on the sidewalks. Section 4 details the implementation and evaluation metrics of the proposed system. Finally, we draw conclusions in Section 5.
\section{2. Related Work}
\label{sec:related}

\subsection{2.1 Predicting Pedestrian Trajectories}

Predicting pedestrian trajectories in urban environments has long been a focal point of research, given its importance in a multitude of applications, ranging from autonomous vehicles to urban planning. In this section, we survey key developments in this area, highlighting the evolution of methodologies and the current state-of-the-art.

\subsubsection{2.1.1 Traditional Methods and Data-Driven Models}
Early approaches to pedestrian trajectory prediction were grounded in physics-based models. The Social Force Model (SFM) introduced by Helbing et al. \cite{helbing1995social} is one such example where pedestrians are modeled as particles and social forces guide their motion. While these models provide intuitive understandings of individual and group behaviors, their deterministic nature limits their capability to capture the intricacies of human movement in dense urban settings. With the surge of data availability, researchers shifted towards data-driven approaches. Gaussian Processes (GPs) were employed by Trautman et al. \cite{trautman2010unfreezing} to predict trajectories by capturing the non-linear relationships between observed pedestrian movements. These methods showed improvements over traditional models but often struggled with computational efficiency in real-time applications.

\subsubsection{2.1.2 Deep Learning Approaches}
The emergence of deep learning techniques has profoundly revolutionized the field of pedestrian trajectory prediction, primarily due to their ability to handle large datasets and model intricate patterns in data.
The initial wave of deep learning models in this domain exploited Recurrent Neural Networks (RNNs) because of their natural ability to handle sequential data. RNNs were primarily employed to capture temporal dependencies in pedestrian movements, providing a more dynamic prediction compared to their static, traditional counterparts. \cite{chung2014empirical}
LSTM networks, an advanced variant of RNNs, soon followed, addressing the vanishing gradient problem typical in conventional RNNs. The memory cells in LSTMs allowed them to remember patterns over longer sequences, making them particularly apt for predicting pedestrian trajectories in diverse settings \cite{vemula2018social}. As previously mentioned, Zhang et al.'s State Refinement for LSTM \cite{zhang2019sr} is a notable mention here, extending LSTMs to account for the social nature of pedestrian interactions.

\subsection{2.2 Risk Assessment and Motion Planning Techniques}
Numerous research and technical approaches have been proposed in the area of motion planning for AVs or robots under different scenarios to enhance intelligence and capability in risk assessment and trajectory generation. The risk assessment module is a prerequisite for motion planning. Risk assessment computes the danger of potential paths of the ego vehicle based on the surrounding environment \cite{kim2017collision}. Since e-Scooter-involved accidents have increased significantly with little information being available regarding the behaviors of on-road e-scooter riders, risk assessment and mitigation of e-scooter crashes is investigated in\cite{prabu2022risk}. Current risk assessment methods are
not designed for multi-vehicle collision scenarios, and the
potential risks from other vehicles and the surrounding
environments are modeled separately \cite{shen2020collision}. To overcome this issue, some researchers have established the model of risks of cars using the potential field method \cite{huang2019motion}. Based on the assessments from the risk evaluation module, the path planner finds the collision-free trajectory that the ego vehicle should follow over the next short or long time period. In \cite{shen2020collision}, a collision mitigation system with features of all-around (360-degree) collision avoidance capability is proposed for automated vehicles on the highway driving scenario. A Nonlinear Model Predictive Control Framework for Whole Body Motion Planning is also explained in \cite{meduri2023biconmp}. As can be seen from the literature, most risk assessments and path planners are all focused on vehicle perspectives, e-Scooter based crash mitigation system has not been studied and investigated extensively and deeply.
\section{3. System Design}
\label{sec:design}

\subsection{3.1 Collision Avoidance System}
This section mainly introduces the basic concept and operation process of the proposed e-Scooter Collision Avoidance System (eCAS) with pedestrians on the sidewalks. The execution conditions and transition of the riding modes of the future automated e-Scooter under control are discussed based on its risk level. The overall architecture for the proposed CAS is also described.

\subsubsection{3.1.1 Basic Concept of eCAS}
As one of the most popular micro-mobility manners, e-scooters are spreading in hundreds of big cities and college towns in the US and worldwide. Meanwhile, e-scooters are also posing new challenges to traffic safety \cite{prabu2022wearable}. This paper proposes an active safety e-Scooter collision avoidance system, which can predict the pedestrians’ trajectories among certain time horizons, evaluate the potential risks associated with surrounding static and moving obstacles, and generate the corresponding collision-free trajectory from the start location to the goal location for e-Scooter. As an operating mode of automated vehicles, CAS has been developed significantly in both highway and rural driving environments, which can eliminate and/or minimize human errors. The development of CAS for e-Scooter under certain riding environments is still an open topic and research areas based on the best of authors’ knowledge due to their specific dynamic behaviors as compared to bicyclists and pedestrians. The proposed eCAS introduced in this paper mainly focuses on specific areas of the sidewalks/bike lanes with pedestrians, which can continuously analyze the driving scene, understand the behavior of other road users, evaluate the risk assessment, and predict its future trajectory. This proactive model is designed to ensure unobstructed movement in areas with substantial pedestrian traffic without collisions.

\subsubsection{3.1.2 Overall System Architecture and Operation Process}
The proposed collision mitigation system for e-Scooter receives the predicted pedestrian moving trajectories from the perception system for all surrounding objects. The overall architecture and operation process of the eCAS are given in Figure 1. We assume that the e-Scooter can continuously monitor the surrounding environment and obtain the required information without any transmission delays or failures, and the e-Scooter non-linear motion dynamics are not considered in this paper. 

\begin{figure}[!htbp]
  \begin{center}
  \includegraphics[height = 1.6in, width = 3.3in]{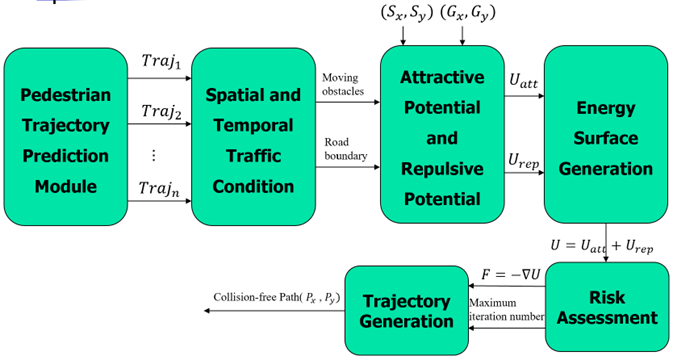}
  \caption{Overall architecture of e-Scooter Collision Avoidance System.}\label{Overall_Architecture}
  \end{center}
\end{figure}

Figure \ref{Overall_Architecture} demonstrates the overall system architecture and operation process of eCAS. As can be seen in the figure, after consuming the predicted pedestrians’ trajectories, the spatial and temporal traffic conditions can be modeled using artificial potential field representation. Then it sends out both the moving obstacles and road boundaries along with the pre-determined eScooter’s start position and goal location to obtain the attractive potential function Uatt and repulsive potential function Urep across the entire interested riding environment. By adding Uatt and Urep, the overall energy surface U is obtained for the further risk assessment module. By taking the negative gradient of the energy surface, the trajectory generation component will generate the optimal collision-free route (X, Y) for e-Scooter considering the maximum iteration number of the gradient descent algorithm, which would push the e-Scooter to move forward while keeping it away from obstacles.
\subsection{3.2 Pedestrian Trajectory Prediction Module}
\begin{figure}[!htbp]
  \begin{center}
  \includegraphics[width = 3.3in]{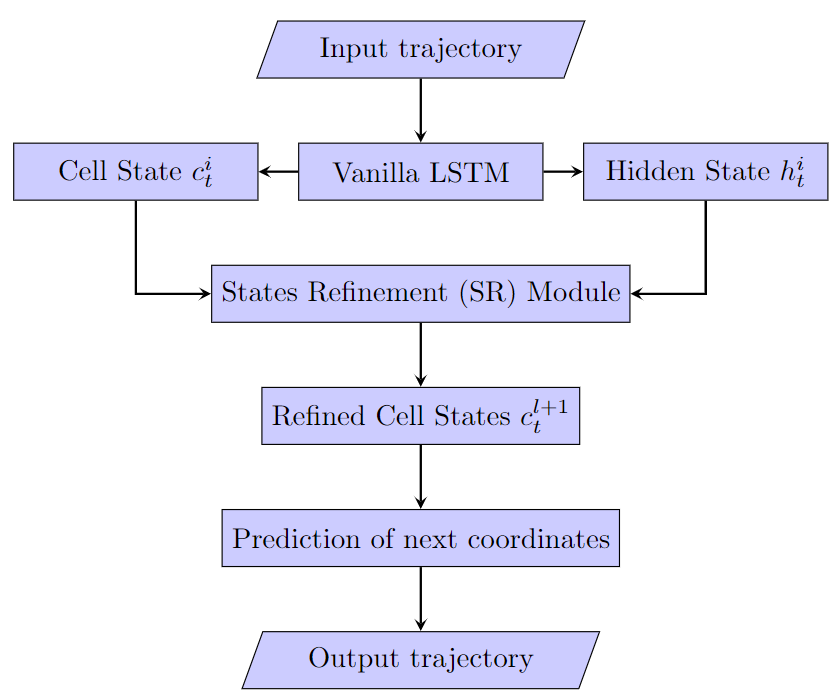}
  \caption{SR-LSTM Trajectory Prediction Flowchar.}\label{SR_LSTM}
  \end{center}
\end{figure}

In the proposed eCAS, we incorporate a sophisticated Pedestrian Trajectory Prediction module based on the States Refinement Long Short-Term Memory (SR-LSTM) framework, as introduced by Zhang et al. \cite{zhang2019sr} as shown in Figure \ref{SR_LSTM} This module is pivotal for understanding and anticipating pedestrian movements, thereby augmenting the predictive capability of eCAS in dynamic environments.

\textbf{Input Trajectory Processing:} The module commences by acquiring the initial trajectory data, which comprises historical pedestrian positions. These positions are crucial for forecasting subsequent movements. The data is first ingested by a Vanilla LSTM layer, which is fundamental in processing sequence information and maintaining dependencies over time. The Vanilla LSTM operates on the input trajectory to produce the preliminary hidden states (\(h_t^i\)) and cell states (\(c_t^i\)), which encapsulate the temporal features of the pedestrian's movement.

\textbf{Hidden and Cell States Generation:} The hidden state (\(h_t^i\)) emerges as the LSTM's memory, holding information pertinent to the pedestrian's prior trajectory. Concurrently, the cell state (\(c_t^i\)) safeguards the long-term context within the LSTM's memory. These initial states are the precursors to the refinement process.

\textbf{States Refinement (SR) Module:} At the core of the SR-LSTM lies the States Refinement (SR) Module, a novel mechanism that enhances the cell states by leveraging the trajectories of neighboring pedestrians and supplementary contextual data. This refinement ensures that the prediction is not solely dependent on the pedestrian's path but is also informed by the surrounding environmental dynamics. Post refinement, the SR Module outputs the refined cell states (\(\hat{c}_t^{l+1}\)), which encapsulate enriched contextual information.

\textbf{Trajectory Prediction:} Leveraging the refined cell states, the module predicts the future coordinates of the pedestrian trajectory. This prediction is rooted in a more sophisticated understanding of the pedestrian's context, thanks to the SR Module's processing.

\textbf{Output Trajectory: } The culmination of this process is the output trajectory, which signifies the eCAS's foresight into the pedestrian's future locations. The refined prediction takes into account not only the pedestrian's past and present positions but also the influence of nearby dynamics.

The SR-LSTM's innovative design, especially the inclusion of the SR module, distinguishes it from conventional LSTM models. By considering a broader range of contextual information, the SR-LSTM significantly elevates the accuracy of trajectory predictions within complex and interactive urban environments, thus enhancing the decision-making capabilities of the eCAS.

\subsection{3.3 Overview of Gradient-based Motion Planner for E-Scooter}
Gradient-based motion planners are widely utilized for robot and automated vehicle local planning based on the Euclidean Signed Distance Field (ESDF), which is significant for modeling the driving environments and evaluating the risks through gradient magnitude and direction. Then a local optimal trajectory can be generated using numerical optimizations on a pre-built ESDF energy surface. As one of the ESDFs, the motion planners relying on artificial potential fields (APF) have been extensively applied in different user cases and proved to be effective for generating collision-free paths in complex environments for practical purposes. Accompanying the convenience of the trips, there have been numerous injuries and fatalities reported in association with micro-mobility options like e-Scooters, according to the estimations from the Centers for Disease Control and Prevention (CDC)\cite{ma2021scooter}. For better handling of the potential collision risks between e-Scooters and pedestrians, this section mainly introduces the overall concept of APF and how to use APF to generate the energy surface for traffic environments. Two types of collision risks are considered in this paper. One is the predicted static and moving pedestrians around the e-Scooter, and the other one is the surrounding driving environment with drivable areas.

\subsubsection{3.3.1 Artificial Potential Field}
The artificial potential field is a motion planning technique for guiding a moving host vehicle from the start position to the goal position in 2-dimensional configuration space by building the artificial potential fields that obey Laplace’s equation, such as gravitational, magnetic, and electrical fields and so on \cite{zhou2020ego}.  The APF algorithm is used to guide the robot to be attracted to the goal and repelled from the obstacles. We can think of the robot as a positively charged object moving toward the negatively charged destination and moving away from the obstacles with the same charge as the robot. Thus, two types of artificial potential fields can be generated in the motion planner, one is the attractive field and the other one is the repulsive field. The path planning algorithm utilizes the APF to regulate the e-Scooter in specific scenarios with pedestrians like university or hotel sidewalks. For our case, a 2-D space to be divided into a grid of cells with obstacles and a goal position will be considered. 

\subsubsection{3.3.2 Energy Surface Generation for Traffic Environments}
\begin{figure}[!htbp]
  \begin{center}
  \includegraphics[height = 2in, width = 2.4in]{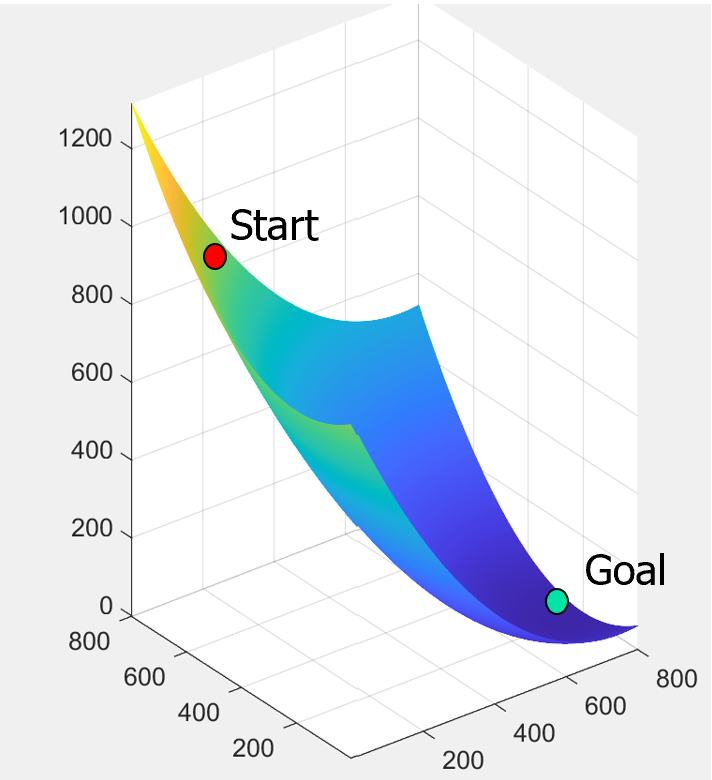}
  \caption{Illustration of the energy surface.}\label{Energy_Surface}
  \end{center}
\end{figure}

With the generation of both attractive fields and repulsive fields based on the goal and obstacles. The potential function of the attractive field and repulsive field can be viewed as the energy surface and the gradient of energy is force. Thus, the energy surface can be modeled to simulate the real driving environment the e-Scooter can be vised as a sphere rolling down the energy surface towards the goal location in a gradient field. As can be seen in Figure \ref{Energy_Surface}, the attractive potential function is the distance from the goal to the start location of e-Scooter with the highest energy away from the start and zero at the goal location. Therefore, based on the energy surface, a gradient-based path planner can be designed using the negative gradient, which is the largest change in energy.
In terms of the function of attraction force, the following formula of distance measure can be utilized to generate the force by the goal location. Thus, $Pos_x$ and $Pos_y$ are the coordinates of the e-Scooter start point, $Pos_gx$ and $Pos_gy$ are the coordinates of the goal node and $h$ is a constant to adjust the strength of the force.
\begin{equation}\label{AttractiveForce}
   F_{att}= h*\sqrt{\left|Pos_x - Pos_{gx}\right|^2 - \left|Pos_y - Pos_{gy}\right|^2}
\end{equation}
Regarding the function of repulsive force, its strength will be increased when the e-Scooter is closer to the obstacles. As described in the previous sections, the drivable areas with boundaries and obstacles are included. The repulsive forces by the road boundaries will remain the same throughout the motion planner due to its uniform geometry, which does not affect the computations afterward. The force generated from boundaries is utilized to keep the e-Scooter riding along the drivable areas.
\begin{equation}\label{BoundaryForce}
    F_{rep} = \frac{1}{\alpha + \sum_{i}^{s}(g_i + \left|g_i\right|)}
\end{equation}
Where $g_i$ is the representation of boundary with a linear function of convex region. $\alpha$ is a small constant value and $s$ is the number of boundary segments.

In the repulsive forces from all obstacles, the following formula can be used to compute the forces.
\begin{equation}
    U_{rep}(q)=\begin{aligned}
        \left\{
    \begin{array}{lr}
    \frac{1}{2\delta}(\frac{1}{D(q)} - \frac{1}{Q^{*}})^{2}, D(q) \leq Q^{*}\\
          0, D(q) > Q^{*}
    \end{array}
        \right.
    \end{aligned}
\end{equation}
Where $\sigma/2$ is the maximum potential, $1/(D(q))$ is the potential of Euclidean distance to the closest obstacle, $Q^*$ is the potential of Euclidean distance around the obstacle.
Hence, the resultant potential function can be represented by adding the attractive potential and repulsive potential together below:
\begin{equation}\label{TotalForce}
    U(q) = U_{att}(q) + U_{rep}(q)
\end{equation}
\subsection{3.4 Motion Planning and Trajectory Generation}
After getting the overall energy surface across the environment including both road boundaries and the predicted pedestrian trajectories, this section will mainly elaborate on how to conduct the risk assessment and generate the safest path through the APF.
\begin{figure*}[ht]
\centering
\includegraphics[width=0.9\textwidth]{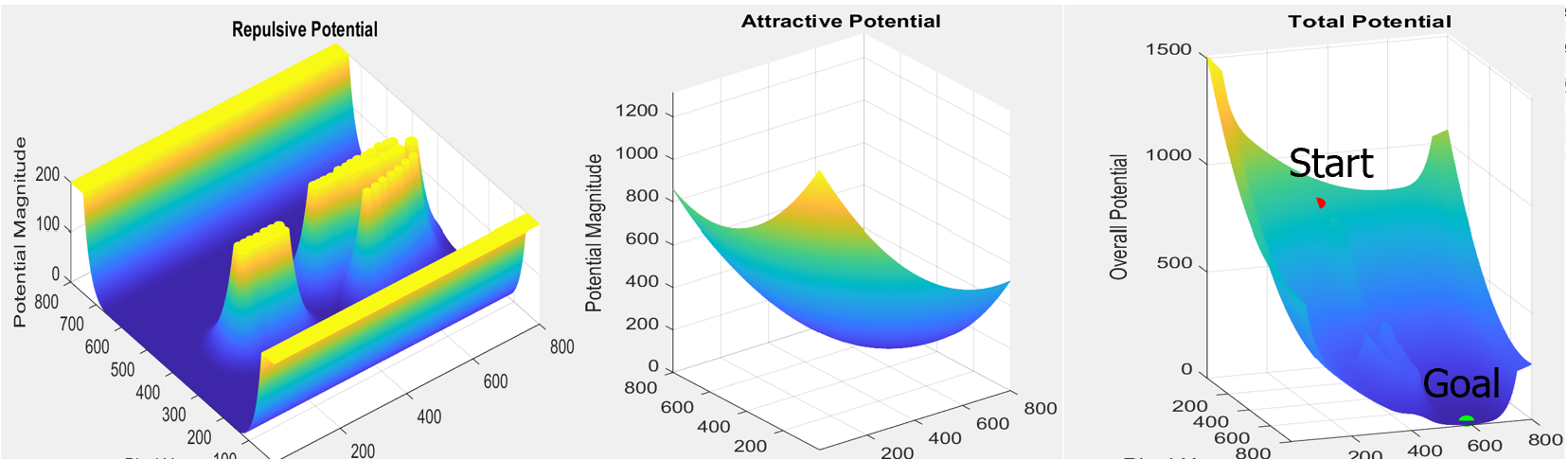}
\caption{Risk Assessment of e-Scooter Riding Environment with the repulsive potential surface, attractive potential surface and total potential surface.}
\label{Risk_Assessment}
\end{figure*}
\subsubsection{3.4.1 Risk Assessment}
In formula (4), the total energy surface has been obtained using a 2D array containing the potential function values across the environment with predicted pedestrians’ trajectories. The start coordinates are also defined as an array specifying the location of the first entry of the x coordinate and y coordinate, as well as the goal coordinate. The maximum number of iterations of the planner that need to be searched is also defined for this specific scenario with e-Scooter and pedestrians.

The risk assessment module is mainly operated through the resultant potential function with attractive potential and repulsive potential, where attractive potential represents the goal position with lower potential values that the e-Scooter would like to reach without collisions after a certain time duration, and the repulsive potential demonstrates and visualize the locations of surrounding obstacles and boundaries with higher potential values. 

As can be seen in Figure \ref{Risk_Assessment}, the potential function according to the e-Scooter riding environment with a red start point and green goal point has been modeled and illustrated from the outputs of the pedestrian trajectory prediction module. The goal destination leads to a lower energy barrier with a smaller potential magnitude in the darker blue color, but the obstacles of pedestrians and road boundaries create a higher energy barrier with a larger potential magnitude in the darker yellow color across the whole riding environment.

\subsubsection{3.4.2 Trajectory Generation}
The safest collision-free trajectory is generated by first computing the gradient of the potential function over the whole e-Scooter riding environment, and then taking the negative gradient which is the largest change in energy surface. Generally, it is known as the gradient descent algorithm which follows a functional surface until you reach its minimum. The negative gradient of the total potential function can be calculated below:
\begin{equation}
    F(q) = -\nabla U(q)
\end{equation}
The detailed procedures for generating the desired trajectory are illustrated below:\\
1.	Determine the start position and goal location of the e-Scooter before motion planning.\\
2.	Receive the predicted pedestrians’ trajectories during a certain time horizon.\\
3.	Generate the resultant potential function for the overall riding environment as well as compute its negative gradient.\\
4.	On every iteration the planner should update the position of the re-Scooter based on the gradient values contained in the arrays $g_x$ and $g_y$ with the normalized gradient vectors.\\
5.	Update the route by adding the new position of the e-Scooter to the end of the route array. Note that the distance between successive locations in the route should not be greater than 1.0.\\
6.	In some corner cases when the planner gets stuck in a local minimum, the motion planner will let the e-Scooter take a random movement by perturbing out of the minima to explore the possible route to reach the goal location.\\
7.	Continue the same procedure until the distance between the robot's current position and the goal is less than 2.0 or the number of iterations exceeds its maximum value.\\
8.	Output the optimal collision-free trajectory with an array with 2 columns (X, Y) that depicts how the e-Scooter position evolves over the state on the gradient of the potential function.\\
\section{4. Experiment}
\label{sec:experiment}
\subsection{4.1 Pedestrian trajectory prediction}
In the evaluation of our trajectory prediction module, we incorporated the ETH \cite{pellegrini2009you} and UCY \cite{lerner2007crowds} datasets, which are comprehensive compilations of pedestrian paths, containing intricate social interactions amongst 1536 individuals. The datasets' extensive collection of non-linear walking trajectories forms the basis for our training regime. To quantify the performance of our model, we used two primary metrics \cite{pellegrini2009you}: the \textbf{Mean Average Displacement (MAD)} error, which is the mean Euclidean distance between the predicted and the actual ground truth positions over all the timesteps, and the \textbf{Final Average Displacement (FAD) }error, which measures the Euclidean distance at the final predicted point, indicating the model's accuracy in determining the pedestrian's endpoint. Following the implementation guidelines detailed by Zhang et al \cite{zhang2019sr}., our model adhered to the prescribed hyperparameters and underwent a rigorous training process of 1000 epochs, processing batches of 64 pedestrians. A specific scenario from the UCY University dataset, chosen for its complexity and the richness of pedestrian interactions, was not only critical for the evaluation of our trajectory prediction module but also served as a basis for subsequent eCAS performance assessments. The results were promising, with the trajectory prediction module achieving a MAD of 0.71 meters and a FAD of 1.46 meters, showcasing a high level of accuracy in forecasting pedestrian movements.

\subsection{4.2 eCAS performance evaluation}
To evaluate and verify the proposed e-Scooter collision avoidance system, some simulation results on two representative scenarios of “parallel moving” and “crossing moving” are illustrated in the following figures.
\begin{figure*}[ht]
\centering
\includegraphics[width=0.9\textwidth]{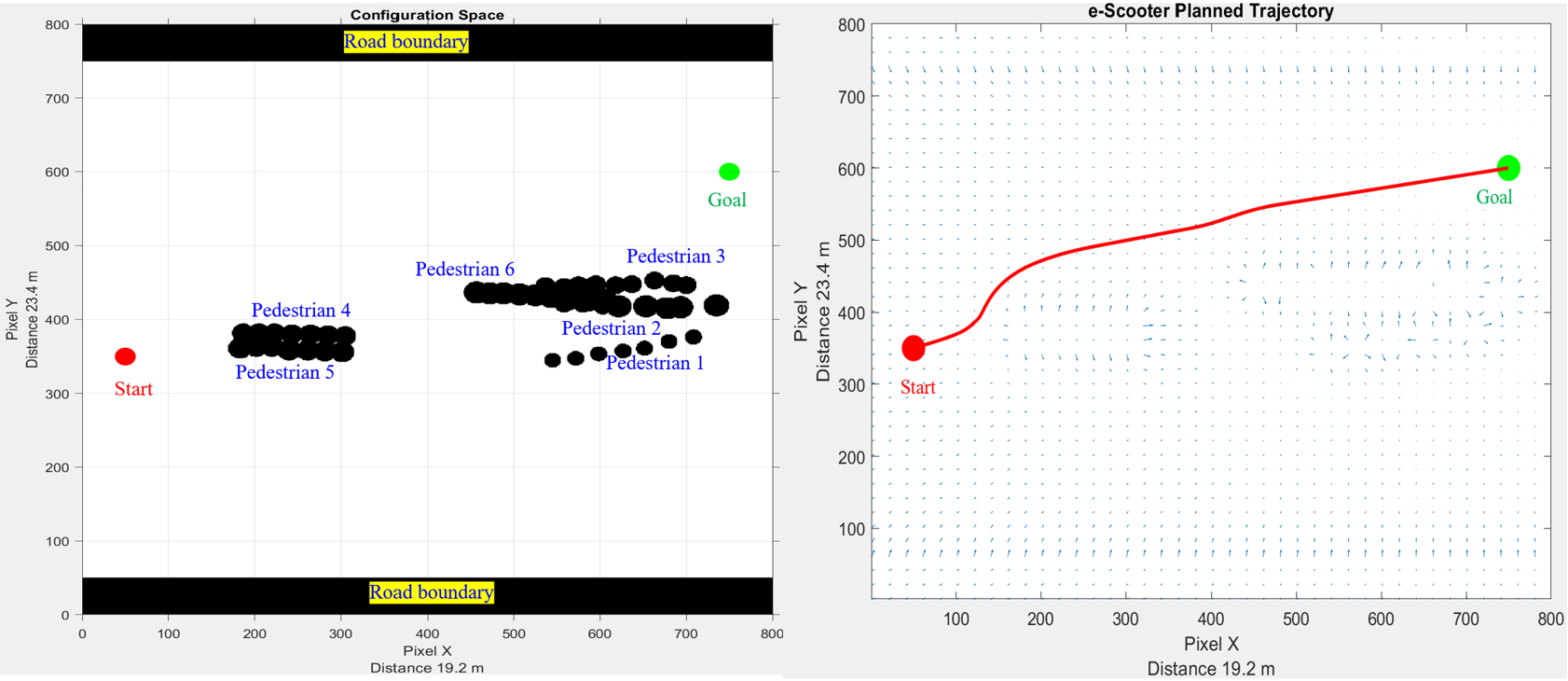}
\caption{e-Scooter riding environment with parallel moving pedestrians and planned trajectory.}
\label{Scenario1}
\end{figure*}
\begin{figure*}[ht]
\centering
\includegraphics[width=0.9\textwidth]{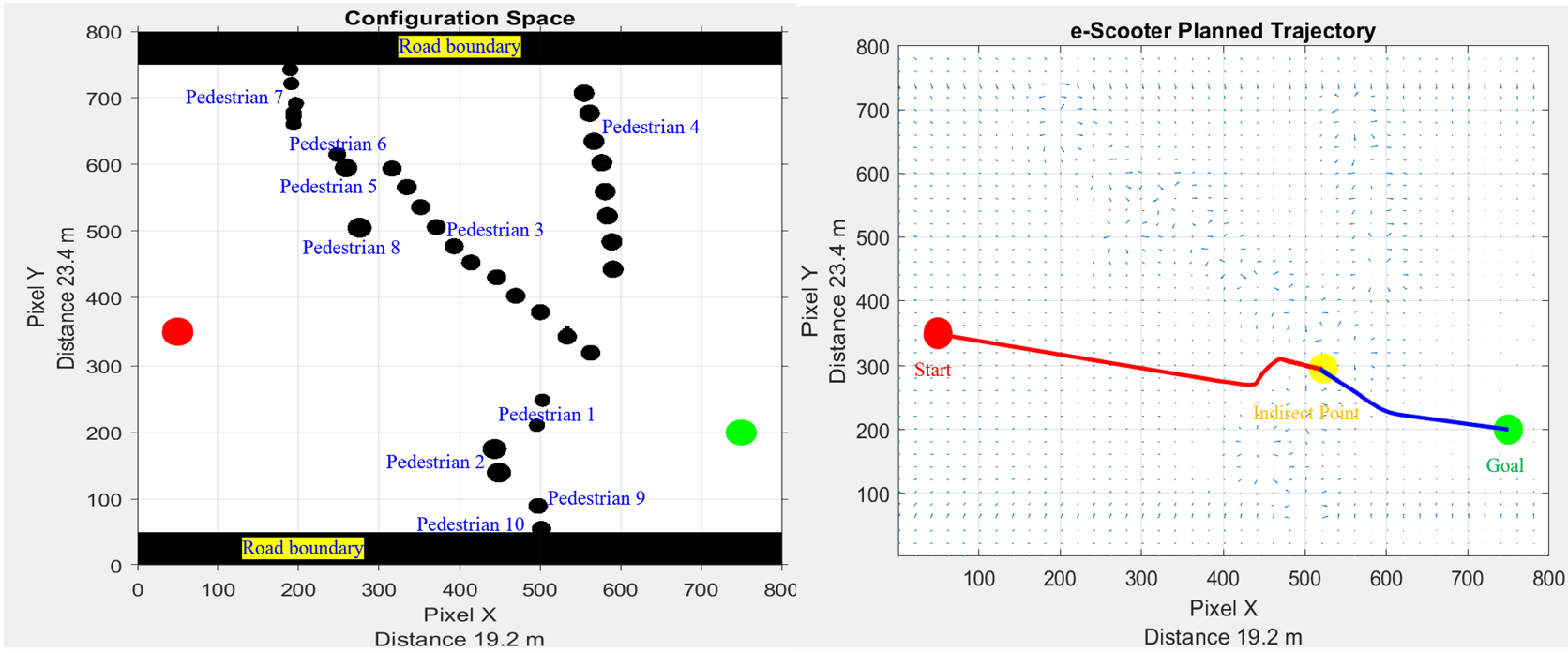}
\caption{e-Scooter riding environment with crossing moving pedestrians and planned trajectory.}
\label{Scenario2}
\end{figure*}
\subsection{Scenario 1 - Parallel Moving}
As shown in Figure \ref{Scenario1} in the “parallel moving” scenario, the e-Scooter will start from the red point and end at a green point with a total of six pedestrians with different velocities and varying moving directions as well as the road boundaries, which has the general parallel motion pattern as compared to the e-Scooter riding direction. To be specific, the prediction model first estimated all six pedestrians’ trajectories for around 4 secs with pedestrian 1 average running speed of 2.79 m/s, pedestrian 2 average walking speed of 0.52 m/s, pedestrian 3 average walking speed of 0.71 m/s, pedestrian 4 average moving speed of 1.97 m/s, pedestrian 5 average walking speed of 1.67 m/s, and pedestrian 6 average moving speed of 1.55 m/s. Then the collision risks around the e-Scooter were evaluated using the energy surface with artificial potential fields. After a gradient decent-based path is planned in eCAS, the e-Scooter’s collision-free trajectory in a global coordinate is generated from the red point to the green point (shown in Figure \ref{Scenario1}).

\subsection{Scenario 2 - Crossing Moving}
In the “crossing moving” scenario from Figure \ref{Scenario2}, the e-Scooter also starts from the red point and ends at a green point but with a total of ten pedestrians with different velocities and varying moving directions as well as the road boundaries. All pedestrians have the general crossing motion pattern as compared to the e-Scooter riding direction, which leads to a more challenging task for e-Scooter to avoid crashes. To be specific, the prediction model first estimated all ten pedestrians’ trajectories for around 4 secs with pedestrian 1 average moving speed of 0.72 m/s, pedestrian 2 average walking speed of 0.73 m/s, pedestrian 3 average walking speed of 0.45 m/s, pedestrian 4 average moving speed of 0.90 m/s, pedestrian 5 average walking speed of 0.17 m/s, pedestrian 6 average moving speed of 0.27 m/s, pedestrian 7 average walking speed of 1.35m/s, pedestrian 8 average moving speed of 0.19m/s, pedestrian 8 average walking speed of 0.46m/s, and pedestrian 10 average moving speed of 0.54m/s. The risk assessment around the e-Scooter was conducted using the energy surface with artificial potential fields. Since a more complex scenario is encountered, the eCAS first locates the collision avoidance path from the start point to the indirect point in yellow in Figure 6, which indicates the planner gets stuck in a local minimum. Then in re-planning, the eCAS guides the e-Scooter to take a random movement by perturbing out of the minima to find the possible optimal route to reach the goal location from the indirect point to the goal location. By combining two planned paths, e-Scooter’s collision-free trajectory in a global coordinate is generated from the red point to the green point (shown in Figure \ref{Scenario2}).
\section{5. Conclusion} \label{sec:conclusion}
This paper proposed a novel e-Scooter collision avoidance system using SR-LSTM-based pedestrian trajectory prediction model and motion planner with the gradient decent algorithm based on the artificial potential field and energy surface. It has been shown that the developed pedestrian trajectory prediction module is able to achieve low Mean Average Displacement (MAD) and Final Average Displacement (FAD) errors, indicating good accuracy in predicting pedestrian paths. From the devised motion planner in eCAS, the scooter-riding environments with both road boundaries and moving pedestrians were modeled and represented using the spatial and temporal space of the potential field with the visualization of risks on an energy surface. The collision-free trajectory with the lowest risk was also generated in the global sense from the start point to the goal location according to the specific test scenarios. Two representative scenarios of "parallel moving" and "crossing moving" were used to verify that the developed eCAS can execute successfully to mitigate the potential crashes between electronic scooters and pedestrians. The simulation results illustrated the effectiveness and capability of the proposed system.

Moving forward, there are several avenues to enhance the effectiveness of e-scooter collision avoidance systems. One area is Sensor Fusion, where integrating a variety of sensors, such as radar and cameras, can improve the model's accuracy and robustness. Another direction is adapting the system for Multimodal Transportation Environments, which requires testing and refining the system's capabilities to ensure smooth interactions between e-scooters, bicycles, cars, and public transportation. Additionally, the system could evolve to consider the impact of the e-scooter's own trajectory on pedestrian behavior. A new model could anticipate how pedestrians might react to the e-scooter's path and adjust in real-time.

\bibliographystyle{ieeetr} 
\bibliography{reference}







\end{document}